\documentclass[conference]{IEEEtran}
\usepackage{cite}
\usepackage{hyperref}
\hypersetup{
	unicode=true,
        bookmarksnumbered=true,
	colorlinks=true,
	linkcolor={red!70!black},
	citecolor={red!70!black},
	urlcolor={blue!70!black},
	pdfborder={0 0 0}
}
\usepackage{url}
\usepackage{amsmath,amssymb,amsfonts}
\usepackage{algorithmic}
\usepackage{graphicx}
\usepackage{textcomp}
\usepackage{xcolor}
\usepackage{xspace}
\usepackage{authblk}
\begin{document}

\newif\ifshowcomment
\showcommenttrue

\ifshowcomment
\newcommand{\todo}[1]{\textcolor{red}{[{TODO: #1}]}}
\newcommand{\pratyush}[1]{\textcolor{orange}{P: #1}}
\newcommand{\theo}[1]{\textcolor{blue}{T: #1}}
\newcommand{\esha}[1]{\textcolor{red}{E: #1}}
\else
\newcommand{\todo}[1]{}
\newcommand{\pratyush}[1]{}
\newcommand{\theo}[1]{}
\newcommand{\esha}[1]{}
\fi

\newcommand{\myparagraph}[1]{\vspace{\medskipamount}\noindent\textit{#1.\xspace}}
\newcommand{\myparagraphnodot}[1]{\vspace{\smallskipamount}\noindent\textit{#1\xspace}}
\newcommand{\myparagraphemph}[1]{\vspace{\smallskipamount}\noindent\emph{#1.\xspace}}
\newcommand{\eg}{\emph{e.g.}\xspace}
\newcommand{\etc}{etc.\@\xspace}
\newcommand{\ie}{\emph{i.e.}\xspace}
\newcommand{\etal}{\emph{et al.}\xspace}
\newcommand*{\rom}[1]{\uppercase\expandafter{\romannumeral #1\relax}}
\newcommand{\insightparagraph}[1]{\vspace{\smallskipamount}\noindent\textbf{\emph{Insight \rom{#1}:\xspace}}}

\title{Input-Dependent Power Usage in GPUs\\
\author{Theo~Gregersen$^1$,
        Pratyush~Patel$^1$,
        Esha~Choukse$^2$%
        \\
        $^1$University of Washington,
        $^2$Microsoft Azure Research - Systems
}


}

\maketitle

\begin{abstract}
GPUs are known to be power-hungry, and due to the boom in artificial intelligence, they are currently the major contributors to the high power demands of upcoming datacenters.
Most GPU usage in these popular workloads consist of large general matrix-matrix multiplications (GEMMs), which have therefore been optimized to achieve high utilization of hardware resources.  

In this work, we show that modifying the input data to GEMMs, while maintaining the matrix shapes and sizes can notably change the power consumption of these kernels.
We experiment with four kinds of input variations: value distribution, bit similarity, placement, and sparsity, across different data types.
Our findings indicate that these variations can change the GPU power usage during GEMM by almost $40\%$.

We hypothesize that input-dependent power usage variations occur due to changes in the number of bit flips in the GPUs.
We propose leveraging this property through compiler and scheduler optimizations to manage power and reduce energy consumption.
\end{abstract}


\section{Introduction}
\label{sec:intro}

Power demand for datacenters, supercomputers, and machine learning is exploding, mainly driven by the growth in large language models (LLMs) \cite{Lin2024Exploding, Wu2021SustainableAE, Lin2023Adapting, Zhao2023NERSC}.
Recent estimates forecast substantial annual increases in datacenter energy consumption and raise concerns about demand exceeding grid capacity in next few years \cite{MckinseyDC, Lin2023Adapting, Blunt2024WSJ}.
With the industry pushing for more compute, addressing power consumption is key to sustainability \cite{Patel2023Agile, Patel2024Characterizing, Patel2023GPU}.
Previous work has explored and utilized various techniques for managing power, such as power capping \cite{Li2020Thunderbolt, Zhang2021Flex},
frequency scaling \cite{Patel2023GPU}, approximation \cite{Mittal2016Approximate, Sampson2011EnerJ}, and batching \cite{You2023Zeus, Chung2023Perseus}.
Our work shows that input data can also be utilized to manage power.

We focus on compute kernels and datatypes applicable to a variety of accelerator workloads.
Accelerators such as GPUs are central to modern machine learning.
For instance, Meta plans to amass around 600,000 H100 GPUs by the end of 2024 \cite{Heath2024Meta}, Microsoft has enabled OpenAI training and inference with large cluster deployments of A100 and H100s\cite{OAItraining}, and the top supercomputers leverage GPUs \cite{top500}.
GPUs are also very power hungry.
For instance, recent NVIDIA H100 SXM5 GPUs have a total power draw of 700 watts \cite{NvidiaH100datasheet}, with a DGX-H100 consisting 8 GPUs needs provisioning of 10,000W\cite{NvidiaH100datasheet}.

Research on reducing accelerator peak power draw or energy often focuses on hardware design and the surrounding infrastructure: system design and scheduling \cite{Choi2023EnvPipe, You2023Zeus, Patel2023Splitwise, 
Chung2023Perseus}, power control \cite{Patel2023GPU, Komoda2013Power}, or efficient chips \cite{Jouppi2018TPU, Qin2020Sigma, Tibaldi2023Survey}.
Instead, we target input data for general matrix multiplication (GEMM) kernels \cite{Cublas, Cutlass, Cusparse}.
GEMM kernels comprise a large portion of machine learning cycles and are important operations for GPUs due to natural compatibility with parallel execution \cite{Qin2020Sigma, NvidiaGEMM, Pati2022BERT}.
Prior work explores efficient implementations of GEMM \cite{Hong2019Adaptive, Gale2020Sparse, Mehrabi2021Sparse} and the performance impacts of quantization \cite{Dettmers2023Spqr, Frantar2023Gptq, Gholami2021Survey}, sparsity, and data ordering \cite{Tang2023TorchSparse} on GEMM in LLMs.
We demonstrate that GEMM input values and placement can have significant impact on GPU power as well.

While research has broadly characterized accelerator power during machine learning and high-performance computing (HPC) workloads \cite{Patel2024Characterizing, Zhao2023NERSC}, minimal prior work measures GPU power consumption due to varying inputs.
Previous work has shown that input data values can significantly impact GPU power, but only considered single instructions such as FMUL or IMUL \cite{lucas2016alupower}.
Bhalachandra et al. investigated the effects of GEMM input patterning on power, but only looked at input value entropy for a single datatype and placement of zero versus non-zero values \cite{Bhalachandra2022UnderstandingTI}.
In addition to input value entropy, we explore the impact of other patterns such as sorting, bit-level sparsity, hamming weight, and similarity.
We also compare across several datatypes and assess the effect of NVIDIA tensor cores.
We find that GEMM input patterns can change GPU power consumption by up to $38\%$.
This observation lends itself to a variety of potential future applications for power and energy efficiency: power-aware sparsity, data pruning for power capping, and efficient data placement algorithms.

\section{Background}

GEMM operations are fundamental to machine learning \cite{sze2017efficientprocessingdeepneural, Pati2022BERT} and many common computational tasks \cite{Lim_2021}.
As such, GEMM is an important target for power efficiency.

GEMM is a fundamental linear algebra operation.
For matrix $A$ with dimensions $(N,K)$, matrix $B$ with dimensions $(K,M)$, matrix $C$ with dimensions $(N,M)$, and scalars $\alpha$ and $\beta$, a standard GEMM execution typically computes the following matrix output \cite{Cutlass}:
$$D = \alpha A \cdot B + \beta C$$

To reduce memory use, the $D$ output matrix is often set to $C$ and updated in-place.
GPU makers such as NVIDIA typically provide proprietary kernels (i.e., compute routines) to execute operations like GEMM on their hardware \cite{NvidiaGEMM}.
Kernel libraries such as cuBLAS \cite{Cublas} and cuSPARSE \cite{Cusparse} are available through public APIs, however the underlying implementations are black boxes.
A more transparent alternative is CUTLASS \cite{Cutlass}, an open-source kernel library maintained by NVIDIA.

To improve performance, NVIDIA GPUs can utilize tensor cores.
Tensor cores are specialized for matrix math operations such as matrix multipy and accumulate (MMA)
and provide acceleration for specific datatypes. 
For instance, the NVIDIA Ampere architecture provides 
20$\times$ FP32 MMA throughput compared to the previous generation \cite{NvidiaA100}.

\section{Experiment Setup}

To explore the impact of GEMM inputs on GPU power, we run a series of GEMM operations on an NVIDIA A100 PCIe virtual machine (VM) hosted on Azure.\footnote{Experiment code and data can be found at \url{https://github.com/theo-gregersen/input-dependent-power-sc24}.}
The NVIDIA A100 PCIe GPU has a maximum thermal design power (TDP) of 300 watts \cite{NvidiaA100}.
We use standard NVIDIA CUTLASS kernels \cite{Cutlass} optimized for GEMM execution, measure power every 100ms with NVIDIA dcgm command-line tools \cite{NvidiaDcgm}, and measure elapsed time with C++ standard library high resolution clocks.

All experiments use 2048 by 2048 matrices.
We selected 2048 as the largest power of two that did not consistently throttle the A100 GPU.
During our experiments, the A100 GPU averaged 98.5\% utilization.
The $C$ matrix is zeroed, and both $A$ and $B$ matrices use the same pattern with $B$ transposed unless otherwise noted.
Reported results are averaged over 10 seeds with 20k iterations each for FP16-T and 10k iterations each for the other datatypes.
The $A$ and $B$ matrices use different seeds.

We explore four datatype setups: 32-bit floating point (FP32), 16-bit floating point (FP16), 16-bit floating point with tensor cores enabled (FP16-T), and 8-bit integer (INT8).
For each datatype, we experiment with value similarity, physical (bit) similarity, data placement, and sparsity.
All of the floating point experiments use the same generated FP32 values, with numeric conversion to their respective datatypes (round to nearest value).
When generating input values, we use appropriate parameters to ensure that all values practically fall within each datatype's representation range.

\begin{figure}[]
    \centering
    \includegraphics[scale=0.6]{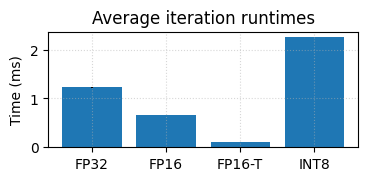}
    \caption{Average iteration runtime by datatype for 2048x2048 GEMM across all A100 GPU experiments. Note that the error bars are a magnitude smaller; iteration runtimes are very consistent across experiments.}
    \label{fig:runtimes}
\end{figure}

\begin{figure}[]
    \centering
    \includegraphics[scale=0.6]{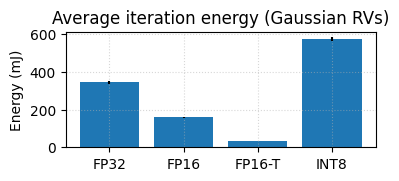}
    \caption{Average iteration energy on A100 GPU for 2048x2048 GEMM filled with random variables from a Gaussian distribution. The distribution has a mean of 0 and a standard deviation of $2^{10}$ for FP and $2^5$ for INT8.}
    \label{fig:energy}
\end{figure}

During testing, we observed slight variations in power measurements depending on when experiments were run.
Power measurements occasionally shifted by up to 10W when the VM instance changed, even when using the same configuration.
We attribute this to process variation across GPUs.
To minimize this effect, we executed all experiments on the same VM instance.
We also trim the first 500ms of power measurements to account for warmup.
Across all experiments for a given datatype, the average iteration runtime (Figure \ref{fig:runtimes}) was consistent to a microsecond-level; this is expected since each experiment uses the standard cutlass kernel.
Figure \ref{fig:energy} shows average iteration energy for a GEMM operation with Gaussian random variables.
Note the identical patterns between the iteration runtimes and energy observations, showing that the power used with random variables is similar across the input types.
In the rest of paper, we report power measurements rather than total energy, as power is the key bottleneck for large-scale machine learning \cite{Lin2024Exploding, Wu2021SustainableAE, Patel2024Characterizing}.

\section{Input-Dependent Power Analysis}
\label{sec:analysis}

We profile different kinds of inputs to GEMM kernels and provide our takeaways (\textbf{Tn}).

\subsection{Value Distribution}

\begin{figure}[]
    \centering
    \includegraphics[scale=0.43]{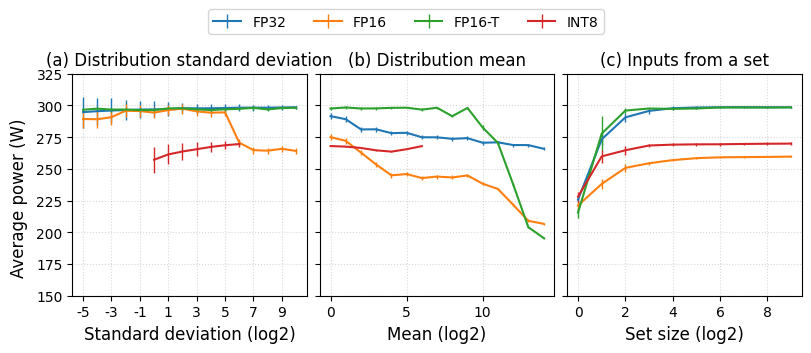}
    \caption{Effects of input value distribution on GPU power.}
    \label{fig:valuedistribution}
\end{figure}

First, we explore the effects of input distribution on GPU power during GEMM.

\textit{Distribution standard deviation}:
Figure \ref{fig:valuedistribution}a shows the average power draw during GEMM when the $A$ and $B$ matrices are filled with Gaussian random variables with a fixed mean of $0$ and varied standard deviation.
\textit{\textbf{T1:} Input distribution standard deviation does not significantly impact power.}

\textit{Distribution mean}:
For the results in Figure \ref{fig:valuedistribution}b, the $A$ and $B$ matrices are filled with Gaussian random variables with a fixed standard deviation of $1$ and varied mean.
\textit{\textbf{T2:} Larger input value means can reduce power for FP datatypes.}

\textit{Inputs from a set}:
The third experiment (Figure \ref{fig:valuedistribution}c) fills $A$ and $B$ with values selected uniformly, with replacement, from a set of Gaussian random variables with a mean of $0$ and standard deviation of $2^{10}$ for FP and $2^{5}$ for INT8.
\textit{\textbf{T3:} Inputs from a small set of unique values decrease power consumption.}

\subsection{Bit Similarity}

\begin{figure}
    \centering
    \includegraphics[scale=0.43]{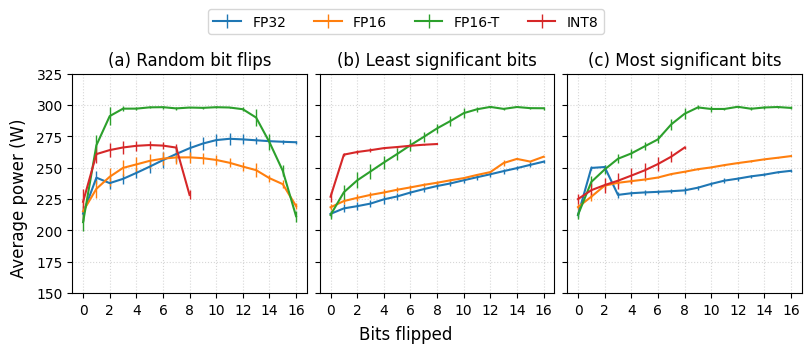}
    \caption{Effects of bit similarity on GPU power.}
    \label{fig:bitsimilarity}
\end{figure}

Next, we consider input data bit similarity. In these experiments, the $A$ matrix is initially filled with one random value and the $B$ matrix is filled with another random value.

\textit{Random bit flips}:
Figure \ref{fig:bitsimilarity}a illustrates how power changes when random bits are flipped in each element.
\textit{\textbf{T4:} Input data with highly similar bits uses less power.}

\textit{Least significant bits}:
Figure \ref{fig:bitsimilarity}b shows how power changes as the least significant bits are randomized.
\textit{\textbf{T5:} As more least significant bits are randomized, power increases.}

\textit{Most significant bits}:
Figure \ref{fig:bitsimilarity}c illustrates how power shifts when the most significant bits are randomized.
\textit{\textbf{T6:} As more of the most significant bits are randomized, power increases.}

\textit{Input data types}: 
Figure~\ref{fig:bitsimilarity} also shows that FP16-T on tensor cores has the highest power usage compared to the other data types. This is important to note, since the default data type in AI applications is FP16-T. Approximation related research tries to reduce the time and memory to run AI inference, but can also have a positive impact on power efficiency.
\textit{\textbf{T7:} FP16-T is the most power hungry data type.}

\subsection{Placement Patterns}

\begin{figure*}
    \centering
    \includegraphics[scale=0.45]{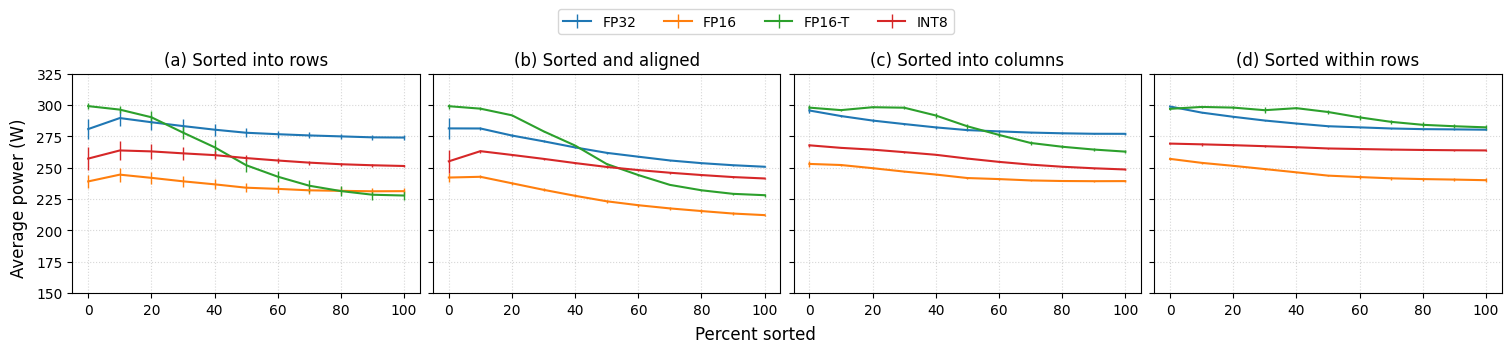}
    \caption{Effects of input value placement on GPU power.}
    \label{fig:placementpatterns}
\end{figure*}

Next, we explore the impact of input data placement on GEMM power.
Across each of the following experiments, the initial values for matrices $A$ and $B$ are constant.
Both matrices are filled with random variables from a Gaussian distribution with a mean of $0$ and standard deviation of $2^{10}$ for FP and $2^{5}$ for INT8.

\textit{Sorted into rows}:
For this experiment (Figure \ref{fig:placementpatterns}a), we partially sort both matrices into rows.
Sorting $n$ percent means that the lowest $n$ percent of values are sorted into the first $n$ percent of indices (row-wise).
The $B$ matrix is not transposed.
\textit{\textbf{T8:} Sorting input values can decrease power consumption.}

\textit{Sorted and aligned}:
For Figure \ref{fig:placementpatterns}b, the matrices are partially sorted into rows again.
However, this time the $B$ matrix is transposed, so the lowest values in $A$ are multiplied with the lowest values in $B$ during GEMM.
\textit{\textbf{T9:} Aligning sorted values decreases power even more than just sorting.}

\textit{Sorted into columns}:
Figure \ref{fig:placementpatterns}c is a similar experiment to that of figure \ref{fig:placementpatterns}a, but the input values are sorted into columns rather than rows.
\textit{\textbf{T10:} Sorting values into columns can decrease power consumption.}

\textit{Sorted within rows}:
We also experiment with sorting within matrix rows and aligning across matrices.
Figure \ref{fig:placementpatterns}d shows how power changes when the $A$ and $B$ matrix rows are partially sorted.
\textit{\textbf{T11:} Intra-row sorting can decrease power, but to a lesser extent than sorting fully.}

\subsection{Sparsity}

\begin{figure*}
    \centering
    \includegraphics[scale=0.45]{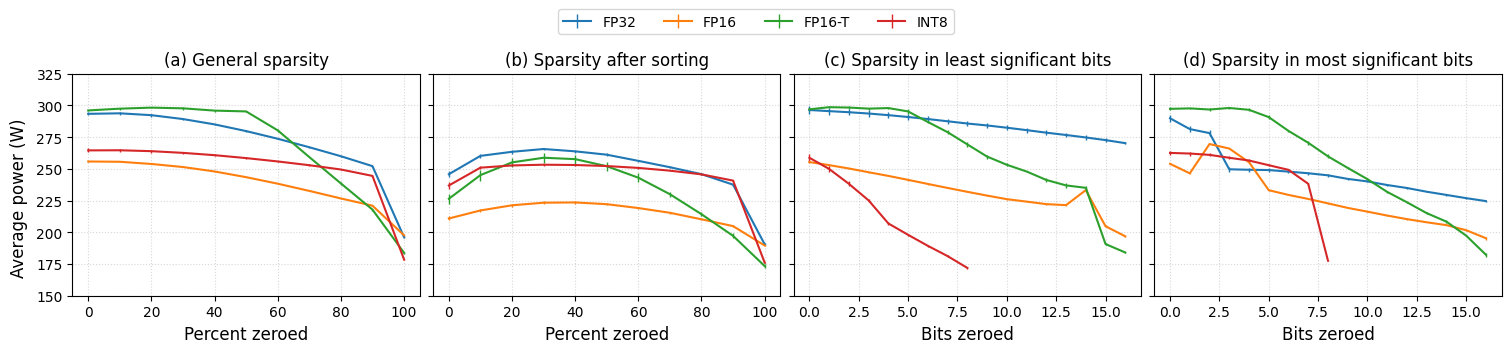}
    \caption{Effects of input value sparsity on GPU power.}
    \label{fig:sparsity}
\end{figure*}

Next, we explore the impact of sparsity on GEMM power.
These experiments use standard GEMM, not sparse GEMM.

\textit{General sparsity}:
Figure \ref{fig:sparsity}a shows power as the matrix is made sparser.
\textit{\textbf{T12:} Matrix sparsity decreases GEMM power.}

\textit{Sparsity after sorting}:
In Figure \ref{fig:sparsity}b, the initial $A$ and $B$ matrices are fully sorted before sparsity is added.
With this patterning, power has a curve that peaks around $30-40\%$ sparsity for floating point datatypes.
Although both sorting and sparsity decrease power in isolation, this trend indicates that they do not compound when paired. 
\textit{\textbf{T13:} Sparsity applied to sorted matrices can actually increase power consumption.}

\textit{Sparsity in least significant bits}:
Finally, we consider sparsity in physical structure.
Figure \ref{fig:sparsity}c is the result of setting each matrix item's least significant bits to zero.
\textit{\textbf{T14:} Zeroing least significant bits can reduce power.}

\textit{Sparsity in most significant bits}:
Figure \ref{fig:sparsity}d illustrates the effect of setting each matrix item's most significant bits to zero.
\textit{\textbf{T15:} Zeroing most significant bits can reduce power.}

\subsection{Generalization}

\begin{figure*}
    \centering
    \includegraphics[scale=0.45]{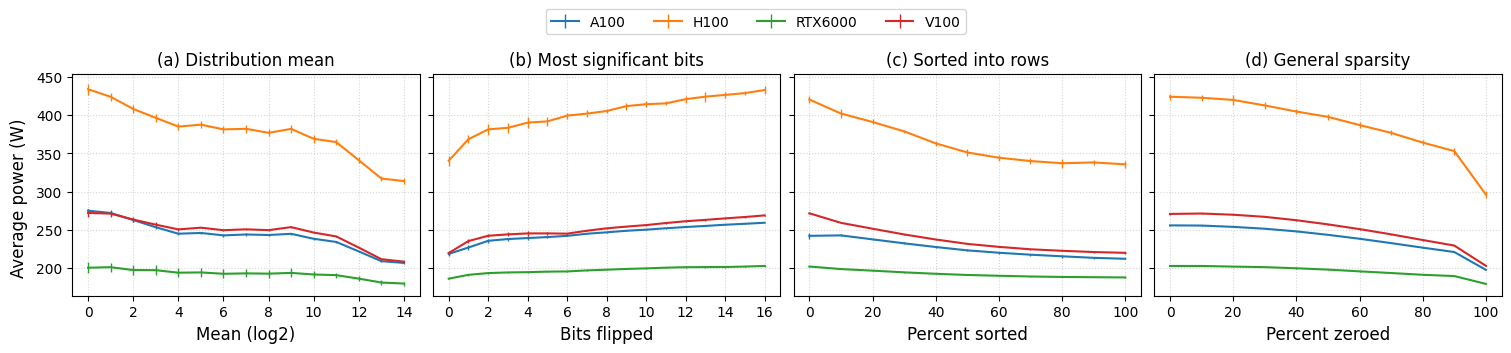}
    \caption{Experiment results across different NVIDIA GPUs.}
    \label{fig:generalization}
\end{figure*}

Our results hold across different GPU generations.
We show this by replicating several of the experiments on an NVIDIA H100 80GB HBM3 GPU (TDP $700$W, local cluster), NVIDIA Quadro RTX 6000 24GB GPU (TDP $260$W, Chameleon cloud \cite{keahey2020lessons}), and NVIDIA Tesla V100-SXM2-32GB GPU (TDP $300$W, 
Chameleon cloud \cite{keahey2020lessons}).
We present results for FP16 runs of the \textit{Distribution mean} experiment, \textit{Most significant bits} experiment, \textit{Sorted into rows} experiment, and \textit{General sparsity} experiment.
Figure \ref{fig:generalization} illustrates the results across GPUs.
The matrix size was $512$ by $512$ for the RTX 6000 (it throttled at $2048$ by $2048$)
For the V100, A100, and H100 GPUs, power consumption trends are consistent.
The RTX 6000 has less prominent changes in power, likely because it is the oldest of the tested GPUs (e.g., uses GDDR6 memory rather than HBM) and has a lower TDP.
\subsection{Bit Alignment and Hamming Weight}

\begin{figure*}
    \centering
    \includegraphics[scale=0.45]{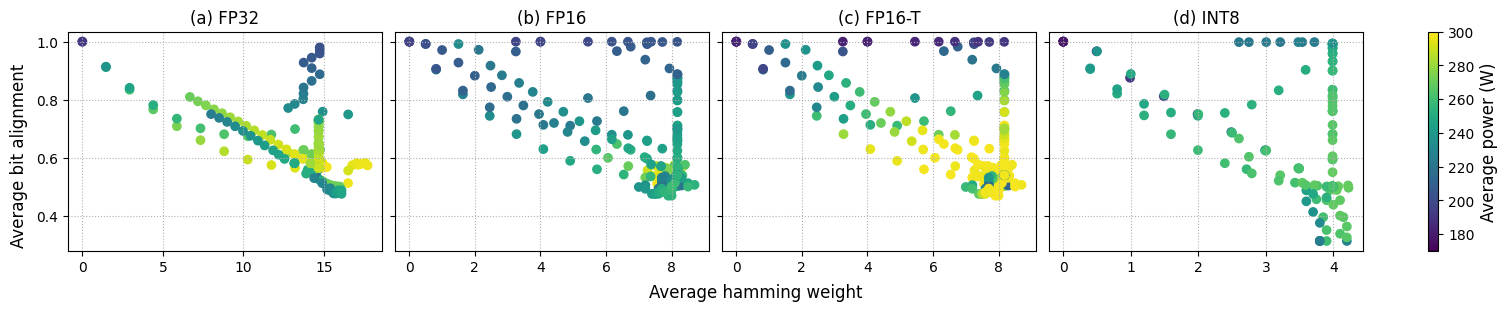}
    \caption{Bit alignment and hamming weight of input values.}
    \label{fig:potential}
\end{figure*}

To investigate broader trends across experiments, we look at bit alignment between values multiplied during GEMM, as well as Hamming weights of the matrix values.
Bit alignment between two values is $0$ if all of the bits are opposite, and alignment is $1$ if all of the bits are the same.
Figure \ref{fig:potential} illustrates average GPU power during GEMM in relation to the average bit alignment between the $A$ and $B$ matrices and average Hamming weight in the $A$ matrix ($B$ has similar weight because of shared patterns).
Each dot represents one experiment configuration from the prior subsections.
Across all floating point datatypes, there seems to be correlation with higher bit alignment or lower Hamming weight and decreasing average GPU power during GEMM.
However, this is not an entirely consistent trend.

\section{Discussion and Future Work}

Going forward, we plan to identify what causes input-dependent power usage variation in GPUs and develop practical techniques to improve the power and energy efficiency of GPU applications at scale.
We list below future directions.

\myparagraph{Identifying Causes}
Based on prior work, we hypothesize that GPU power draw depends on inputs due to changes in the number of bitflips during computation~\cite{pekhimenko2016case}, or how many bits are set \cite{lucas2016alupower}. 
For example, running a GEMM with zero matrices would incur no bitflips, and thus, it likely has lower power draw.
Value similarity likely helps by reducing bit flipping at the hardware level.
We plan to do extensive experiments to validate this hypothesis and investigate other hardware-level factors that might contribute to reduced power draw.

\myparagraph{Input-dependent GPU Power Models}
We are building input-dependent GPU power models to more precisely capture how input variations impact the GPU peak power draw.
Such a power model would take in different data patterns as inputs (\eg, specified via a domain-specific language), and estimate the power usage as output.
Using such power models, future work could build power-aware compilers and optimizers to reduce the power draw of GPU applications that can tolerate input variations.

\myparagraph{Power- and Energy-efficient Machine Learning}
Since machine learning applications, like large language models, are very power intensive~\cite{Patel2024Characterizing}, a key future direction is to leverage input changes to drive down their power and energy usage.
Specifically, we are exploring three different directions.
First, we are trying to modify model weights into value ranges that use less power; for example, shifting the weight values towards larger averages could reduce the power draw as shown in Section~\ref{sec:analysis}.
Second, we plan to investigate whether we can partially or fully sort neural network model weights so as to reduce power draw. 
Since weights within a layer correspond to independent neurons, rearrangement is computationally equivalent as long as each neuron processes its own input.
Recent work leverages permutation invariant transformations to manipulate GPU tiles without changing the computation results \cite{zheng2023pit}.
Similar transformations could potentially be used to set patterns that reduce power.
Finally, we would like to develop sparsity designs that reduce power usage while also optimizing performance, accuracy, and/or memory trade-offs~\cite{Tang2023TorchSparse}.
While it is challenging to estimate/limit the impact of input variations on model accuracy, we are hopeful that the benefits will outweigh the pitfalls.

\ifCLASSOPTIONcompsoc
  \section*{Acknowledgments}
\else
  \section*{Acknowledgment}
\fi
This work is supported by NSF CNS-2104548 and the UW Center for the Future of Cloud Infrastructure (FOCI).
Some results in this paper were obtained using the Chameleon testbed supported by the National Science Foundation.

\bibliographystyle{ieeetr}
\bibliography{paper}

\end{document}